\documentclass[]{spie}

\usepackage{amsmath,amsfonts,amssymb}
\usepackage{graphicx}
\usepackage{booktabs}
\usepackage[colorlinks=true, allcolors=blue]{hyperref}
\usepackage[table]{xcolor}
\usepackage{bbm}
\usepackage{fontawesome5}

\title{Center-Aware Ordinal Learning for Prostate Cancer Grading from Histopathology Images}

\author[1,3]{Mohammad Mahdi Abootorabi}
\author[2,3]{Parvin Mousavi}
\author[1]{Purang Abolmaesumi}
\affil[1]{The University of British Columbia, Vancouver, BC, Canada}
\affil[2]{Queen's University, Kingston, ON, Canada}
\affil[3]{Vector Institute, Toronto, ON, Canada}

\begin{document}
\maketitle

\begin{center}
\faGithub\ Code: \url{https://github.com/DeepRCL/ordinal-multicenter-representation-learning}
\end{center}

\begin{abstract}
Prostate cancer grading from histopathology whole-slide images (WSIs) is complicated by the ordered nature of ISUP grades, class imbalance, and center-dependent variations in staining, tissue preparation, and scanning. Most weakly supervised methods treat grade groups as independent categories and may therefore learn representations that do not transfer reliably across institutions. We propose a center-aware ordinal learning framework that combines a pretrained pathology encoder with attention-based multiple instance learning. The framework uses an Earth Mover's Distance loss to account for ordinal errors, an ordinal supervised contrastive objective to align same-grade slides across centers while separating grades according to disease severity, and curriculum learning to reduce the dominance of benign cases. We evaluate the framework on approximately 9,100 biopsy WSIs from the two-center PANDA dataset. The proposed framework achieves an accuracy of 0.842 and a quadratic weighted kappa of 0.958 on the held-out test set. Its main benefit is observed in cross-center transfer: when a linear probe is trained on Karolinska representations and evaluated on Radboud, accuracy increases from 0.643 to 0.823 and quadratic weighted kappa from 0.899 to 0.930. Performance remains stable in the reverse direction. UMAP projections further show improved mixing between centers while preserving an ordinal grade trajectory. These results suggest that combining ordinal and center-aware objectives produces more transferable slide representations for multi-center prostate cancer grading.
\end{abstract}


\section{Introduction and Related Work}
\label{sec:intro}
Prostate cancer is among the most commonly diagnosed cancers in men worldwide, and histopathology grading remains central to risk stratification and treatment planning \cite{sung2026global,leslie2024prostate}. We formulate this grading problem as a \textit{slide-level ordinal classification} task, where each whole-slide image (WSI) receives a label in $\{0,\ldots,5\}$; 0 denotes benign tissue, while 1--5 correspond to the ISUP Grade Groups. Weakly supervised multiple instance learning (MIL) approaches enable WSI-level learning from slide labels without requiring dense annotations \cite{campanella2019clinical}. In prostate cancer, the PANDA challenge demonstrated that automated grading systems can reach pathologist-level agreement on independent cohorts \cite{bulten2022artificial}. Subsequent studies have explored graph-based weakly supervised grading \cite{xiang2023automatic} and Gleason pattern-aware MIL with uncertainty regularization \cite{hao2026dual}.

Despite these advancements, most WSI grading methods treat grade groups as independent categories and do not explicitly account for their clinical order. Ordinal objectives can reduce severe classification errors by assigning larger penalties to predictions farther from the true grade \cite{litvinov2026clinically}. Generalization across medical centers is another challenge, as differences in tissue preparation, staining protocols, and scanner hardware can introduce center-specific patterns \cite{asadi2024learning}. The imbalanced distribution of grade groups further biases training toward common classes, particularly in prostate data where benign tissue typically dominates. Although these challenges have been studied separately, their joint treatment in prostate cancer grading remains limited.

We propose a center-aware ordinal learning framework built on attention-based MIL. Our method introduces a supervised contrastive learning objective that up-weights cross-center pairs to enforce domain invariance and learn center-agnostic features. To model clinical grade order, we replace the cross-entropy classification loss with Earth Mover’s Distance and add a center-aware ordinal supervised contrastive objective. This is coupled with curriculum learning to mitigate the dominance of benign cases. On the two-center PANDA dataset, the proposed model improves Karolinska-to-Radboud linear-probe accuracy from 0.643 to 0.823 and quadratic weighted kappa (QWK) from 0.899 to 0.930, while maintaining performance in the reverse direction. The model achieves comparable overall performance while achieving a more transferable representation space. These results suggest that jointly modeling grade order and center variation produces more transferable slide representations.

\begin{figure}[t]
    \centering
    \includegraphics[width=\textwidth]{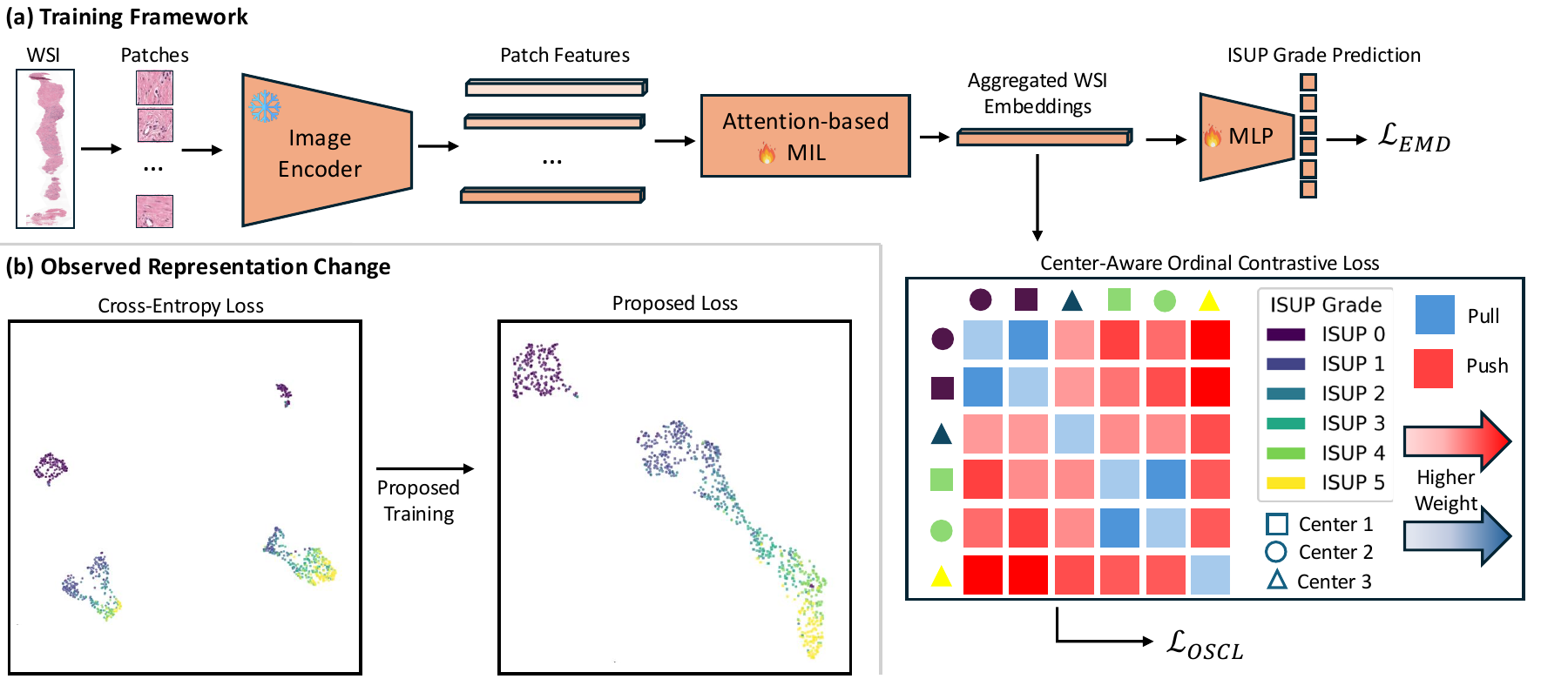}
    \caption{Overview of the proposed framework and learned slide representations. \textbf{(a)} Attention-based MIL aggregates frozen patch features into slide-level embeddings. These are optimized via squared Earth Mover's Distance ($\mathcal{L}_{EMD}$) and center-aware ordinal supervised contrastive loss ($\mathcal{L}_{OSCL}$). In the weight matrix, color denotes ISUP grade and shape denotes center. $\mathcal{L}_{OSCL}$ pulls same-grade pairs together and pushes differing grades apart, up-weighting cross-center pairs and greater ordinal distances. \textbf{(b)} UMAP projections show the proposed method reduces center separation and forms a clearer ordinal trajectory compared to standard cross-entropy.}
    \label{fig:pipeline}
\end{figure}

\section{Materials and Methods}
\label{sec:methods}
Figure~\ref{fig:pipeline} summarizes the proposed training framework and illustrates how the center-aware ordinal objective changes the structure of the learned slide representations.

\subsection{Dataset and Feature Extraction}
We use the publicly available PANDA dataset \cite{bulten2022artificial}, which contains H\&E-stained prostate biopsy WSIs from Karolinska Institute, Sweden, and Radboud University Medical Center, the Netherlands. The cleaned metadata contains 9,128 unique WSIs, including 5,121 from Karolinska and 4,007 from Radboud. Each WSI is assigned a label in $\{0,\ldots,5\}$, where 0 denotes benign tissue and 1--5 correspond to the ISUP Grade Groups. The class distribution is imbalanced, with 2,758, 2,438, 1,133, 825, 984, and 990 slides in grades 0--5, respectively.

Each WSI is divided into non-overlapping $256\times256$ pixel patches at $20\times$ magnification. Patch features are extracted using the frozen UNI2-h pathology foundation model \cite{chen2024uni}, producing a 1536-dimensional representation for each patch. A slide containing $N$ patches is therefore represented as a bag $\mathbf{X}\in\mathbb{R}^{N\times1536}$.

\subsection{Attention-Based Multiple Instance Learning}
We use gated attention-based multiple instance learning (ABMIL) \cite{ilse2018attention} to aggregate the variable-length patch bags into a 768-dimensional slide representation. The slide representation is passed to a two-layer classification head, $768\rightarrow384\rightarrow6$, to predict the ISUP grade. The UNI2-h encoder remains frozen, while the MIL aggregation and classification layers are optimized during training.

\subsection{Center-Aware Ordinal Learning}
We replace the standard cross-entropy classification loss with squared Earth Mover's Distance (EMD) and use center-aware ordinal supervised contrastive learning as an auxiliary objective:
\begin{equation}
\mathcal{L}
=
\mathcal{L}_{\mathrm{EMD}}
+
\lambda\mathcal{L}_{\mathrm{OSCL}}.
\end{equation}

For a batch of $B$ slides and $K=6$ ordered classes, the EMD loss is:
\begin{equation}
\mathcal{L}_{\mathrm{EMD}}
=
\frac{1}{B}
\sum_{b=1}^{B}
\sum_{k=1}^{K}
\left(
\sum_{i=1}^{k}p_{b,i}
-
\sum_{i=1}^{k}t_{b,i}
\right)^2,
\end{equation}
where $p_{b,i}$ and $t_{b,i}$ denote the predicted probability and one-hot target for class $i$, respectively. This loss assigns larger penalties to predictions farther from the true grade \cite{hou2016squared}.
Standard supervised contrastive learning pulls samples with the same label together and separates samples with different labels \cite{khosla2020supervised}. We extend this formulation using both grade distance and center identity. Slides with the same grade form positive pairs, with cross-center positive pairs receiving greater weight. Different-grade slides form negative pairs whose weights increase with their ordinal distance, and cross-center negative pairs are further up-weighted. Specifically, the positive and negative pair weights are:
\begin{equation}
w^{+}_{ij}
=
1+\lambda_d\mathbbm{1}[c_i\neq c_j],
\qquad
w^{-}_{ij}
=
\left(1+\lambda_o\frac{|g_i-g_j|}{K-1}\right)
\left(1+\lambda_d\mathbbm{1}[c_i\neq c_j]\right),
\end{equation}
where $g_i$ and $c_i$ denote the grade and center of slide $i$. We use $\lambda_o=\lambda_d=1$. In the contrastive objective, the negative weights scale the exponentiated negative similarities in the denominator, whereas the positive weights determine the weighted average over same-grade pairs. Each anchor is further weighted by the product of a hardness term, $1+(1-p_{i,g_i})$, and the inverse frequency of its grade in the training set. These anchor weights are normalized within each batch.

\paragraph{Curriculum Weighting.} We introduce class-dependent weighting after an initial three-epoch warm-up. Lower-grade classes receive progressively smaller weights, with the largest reduction applied to benign slides, while higher-grade cases retain greater influence on the EMD objective. This shifts training toward underrepresented and more difficult cancer grades without discarding benign samples.

\subsection{Baselines and Experimental Setup}
The proposed method and its ablations are trained for 40 epochs using AdamW with a learning rate of $8\times10^{-4}$, weight decay of $10^{-4}$, batch size 16, and cosine learning-rate decay. The base contrastive coefficient is 0.5 and is adapted using an exponential moving average of the unweighted auxiliary-to-total loss ratio, with a target ratio of 0.05. We compare the proposed framework with: (i) a cross-entropy baseline, (ii) focal loss with $\gamma=2$ \cite{lin2017focal}, (iii) EMD ordinal loss alone, (iv) cross-entropy with supervised contrastive learning, and (v) ablations that remove center-aware weighting or curriculum learning.
Performance is reported using accuracy, macro-averaged F1 score, and quadratic weighted kappa (QWK). To evaluate representation transfer, we freeze the learned slide embeddings and train an $\ell_2$-regularized multinomial logistic regression probe on one center before evaluating it on the other. We report both Karolinska-to-Radboud and Radboud-to-Karolinska transfer. Finally, we project the test embeddings into two dimensions using UMAP \cite{mcinnes2018umap}.

\section{Results and Discussion}
\label{sec:results}

\subsection{Grading Performance and Ablation Study}

Table~\ref{tab:ablations} compares the classification objectives and the components of the proposed framework with baselines. Focal loss achieved the highest accuracy of 0.866, whereas EMD produced the highest QWK of 0.964. Focal loss and EMD both achieved a macro-F1 score of 0.835 at the reported precision. The higher ordinal agreement obtained with EMD supports its use as the main classification objective.

Adding contrastive learning and the center-aware and curriculum components did not improve the overall test-set metrics beyond EMD alone. The proposed model achieved an accuracy of 0.842, macro-F1 of 0.809, and QWK of 0.958. However, adding center-aware weighting to EMD, OSCL, and curriculum learning increased accuracy from 0.829 to 0.842 and macro-F1 from 0.791 to 0.809. These results indicate that the proposed framework achieves overall performance comparable to other loss functions. The ablations show that center-aware weighting partially recovers the grading performance lost after adding OSCL and curriculum learning, while the main benefit of the complete framework is observed in cross-center transfer.

\begin{table}[t!]
\centering
\caption{Comparison of baseline loss functions and ablation study of the proposed loss components on the PANDA test set. OSCL denotes Ordinal Supervised Contrastive Loss, Center denotes center-aware contrastive weighting, and Curr.\ denotes curriculum weighting.}
\label{tab:ablations}
\small
\setlength{\tabcolsep}{6pt}
\begin{tabular}{lccc|ccc}
\toprule
\textbf{Loss Configuration} & \textbf{OSCL} & \textbf{Center} & \textbf{Curr.} & \textbf{Accuracy} & \textbf{Macro-F1} & \textbf{QWK} \\
\midrule
CE                         & -- & -- & -- & 0.824 & 0.784 & 0.956 \\
Focal                      & -- & -- & -- & \textbf{0.866} & \textbf{0.835} & 0.959 \\
EMD                        & -- & -- & -- & 0.865 & \textbf{0.835} & \textbf{0.964} \\
\midrule
CE + OSCL                  & \checkmark & -- & -- & 0.860 & 0.825 & 0.963 \\
EMD + OSCL                 & \checkmark & -- & -- & 0.838 & 0.798 & 0.958 \\
EMD + OSCL + Curr.         & \checkmark & -- & \checkmark & 0.829 & 0.791 & 0.953 \\
EMD + OSCL + Center        & \checkmark & \checkmark & -- & 0.843 & 0.805 & 0.957 \\
\rowcolor{gray!15} \textbf{Proposed} & \checkmark & \checkmark & \checkmark & 0.842 & 0.809 & 0.958 \\
\bottomrule
\end{tabular}
\end{table}

\subsection{Cross-Center Representation Transfer}
We next evaluate whether the learned slide representations transfer between centers. A linear probe was fitted to frozen training embeddings from one center and evaluated on test embeddings from the other. The proposed framework provided its largest improvement in the Karolinska-to-Radboud direction, increasing accuracy from 0.643 to 0.823 and QWK from 0.899 to 0.930 compared with the CE baseline. In the reverse direction, accuracy increased slightly from 0.811 to 0.815, while QWK improved from 0.916 to 0.939.

The asymmetric results suggest that the magnitude of the domain shift depends on the transfer direction. Nevertheless, the proposed method improves metrics in both directions, which indicates that the center-aware ordinal objective produces representations that are less dependent on the source center artifacts.

\begin{table}[t]
\centering
\caption{Cross-center linear probing of frozen slide embeddings. An $\ell_2$-regularized multinomial logistic-regression probe is fitted to training embeddings from one center and evaluated on test embeddings from the other. K and R denote Karolinska and Radboud, respectively.}
\label{tab:linear_probe}
\small
\begin{tabular}{llcc}
\toprule
\textbf{Representation} & \textbf{Transfer} &
\textbf{Accuracy} & \textbf{QWK} \\
\midrule
CE baseline & K $\rightarrow$ R & 0.643 & 0.899 \\
Proposed Method  & K $\rightarrow$ R & \textbf{0.823} & \textbf{0.930} \\
\midrule
CE baseline & R $\rightarrow$ K & 0.811 & 0.916 \\
Proposed Method & R $\rightarrow$ K & \textbf{0.815} & \textbf{0.939} \\
\bottomrule
\end{tabular}
\end{table}

\subsection{Representation Visualization}
\begin{figure}[t]
\centering
\includegraphics[width=0.95\textwidth]{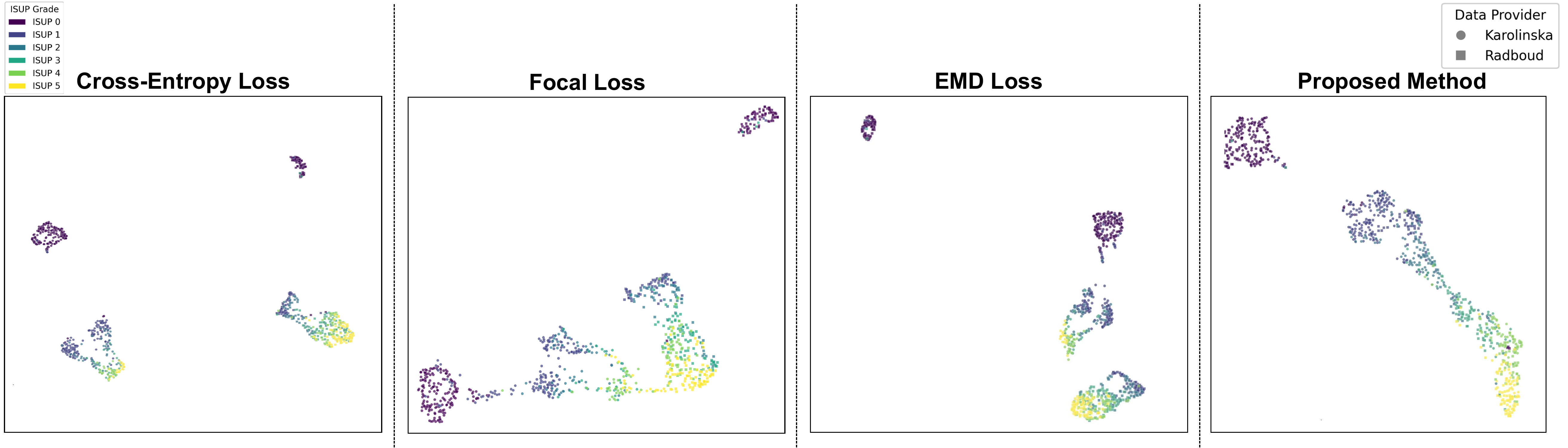}
\caption{UMAP projections of test-set slide embeddings colored by ISUP grade for CE, focal loss, EMD, and the proposed loss. The proposed method produces a more continuous progression from ISUP 0 to 5 and shows greater mixing between the two centers. UMAP is used for qualitative comparison of the representation structure.}
\label{fig:umap_isup}
\end{figure}

\begin{figure}[t]
\centering
\includegraphics[width=0.70\textwidth]{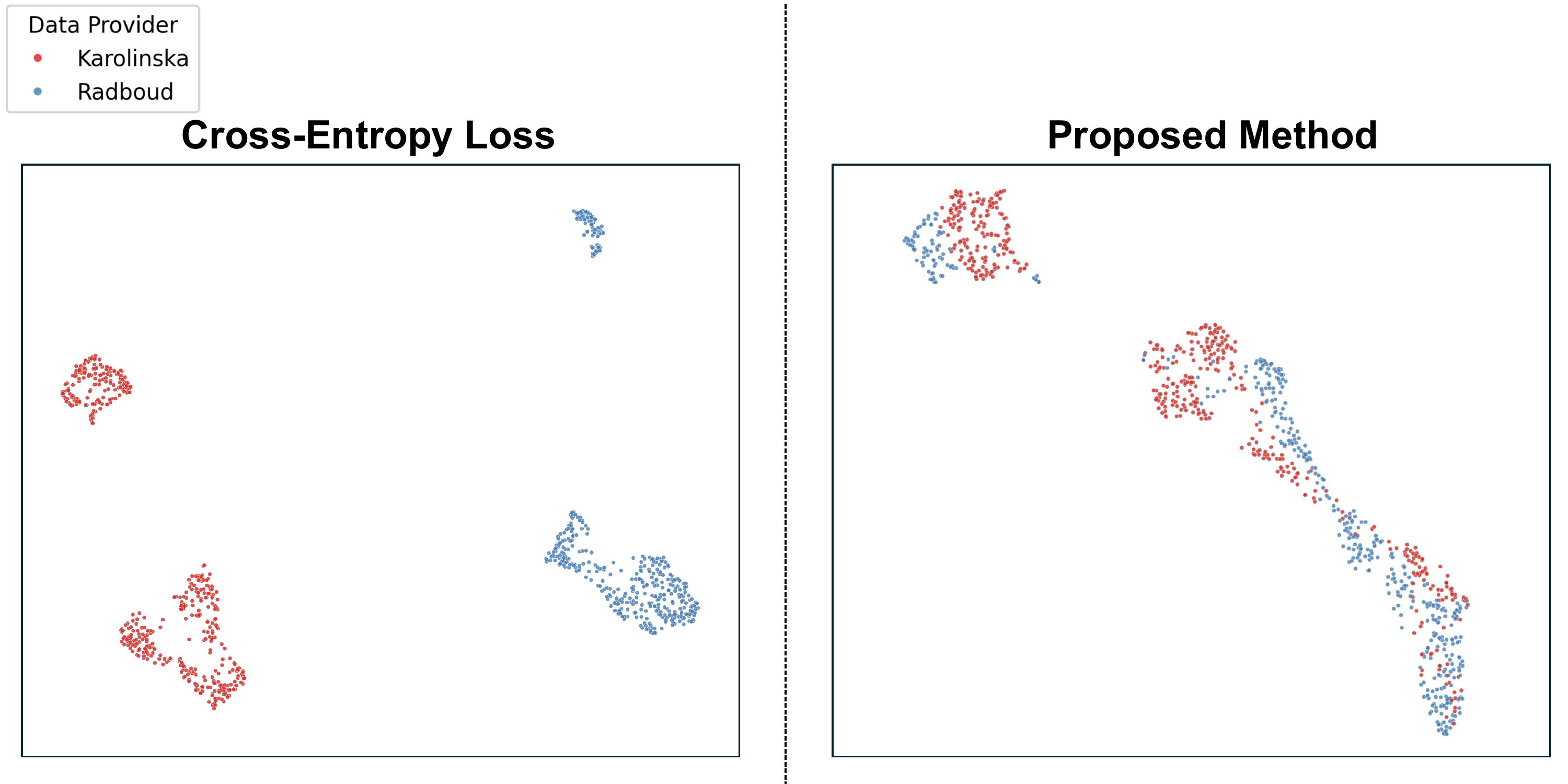}
\caption{UMAP projections of test-set slide embeddings colored by data provider for the CE baseline (left) and proposed method (right). The CE embeddings separate strongly by center, whereas the proposed framework shows greater mixing between Karolinska and Radboud along a shared representation manifold.}
\label{fig:umap_provider}
\end{figure}

Figure~\ref{fig:umap_isup} compares the grade structure of the learned embeddings. The CE and focal-loss representations form several separated clusters without a consistent grade progression. EMD produces a clearer ordering of the grades, but the representation remains divided into separate trajectories. The proposed method forms a more continuous transition from benign tissue to higher-grade disease while retaining separation between neighboring grade groups. When the same embeddings are colored by center, the CE representation shows strong separation between Karolinska and Radboud (Figure~\ref{fig:umap_provider}). In contrast, the proposed method shows substantially greater mixing of the two centers along a shared manifold. This qualitative result is consistent with the cross-center probing results.

\paragraph{Discussion}
The experiments show a trade-off between pooled grading performance and representation transfer. EMD alone achieved the strongest QWK on the combined test set, while the complete framework produced the largest improvement in cross-center linear probing. This distinction is important because strong performance on a mixed-center test set does not necessarily imply that the learned representation transfers reliably between institutions and centers without retraining. The center-aware objective aligns same-grade slides from different centers while preserving separation according to grade distance. The probing and UMAP results suggest that this reduces center-associated structure without removing the ordinal organization needed for grading.

\section{Conclusion}
\label{sec:conclusion}
We presented a center-aware ordinal learning framework for prostate cancer grading from WSIs. While the proposed method achieves performance comparable to baseline loss configurations, it demonstrates superior robustness by yielding highly transferable slide representations. Our results suggest that jointly modeling clinical grade order and center-specific variations improves cross-domain transferability without significantly compromising overall grading accuracy. Future work will explore test-time adaptation strategies to dynamically align out-of-distribution slides at inference time.

\acknowledgments
This work was supported in part by the Canadian Institutes of Health Research (CIHR), the Natural Sciences and Engineering Research Council of Canada (NSERC), Vector Institute, and through computational resources and services provided by Advanced Research Computing at the University of British Columbia. P. Mousavi is supported in part by Canada CIFAR AI Chair and Canada Research Chair. We also thank Sogand Golshahian for her assistance with this work.

\bibliography{report}
\bibliographystyle{spiebib}

\end{document}